\documentclass[journal]{IEEEtai}

\usepackage[colorlinks,urlcolor=blue,linkcolor=blue,citecolor=blue]{hyperref}

\usepackage{color,array}
\usepackage{cite}
\usepackage{amsmath,amssymb,amsfonts}
\usepackage{mathtools, bm}
\usepackage{algorithmic}
\usepackage{subcaption}
\usepackage{graphicx}
\newtheorem{theorem}{Theorem}[section]
\newtheorem{corollary}{Corollary}[theorem]
\newtheorem{lemma}[theorem]{Lemma}

\DeclareMathOperator*{\argmin}{arg\,min}
\newtheorem{prop}{Proposition}
\def\vx{{\bm{x}}}
\def\vF{{\bm{F}}}
\def\cL{{\mathcal{L}}}
\def\cN{{\mathcal{N}}}
\usepackage{textcomp}
\def\BibTeX{{\rm B\kern-.05em{\sc i\kern-.025em b}\kern-.08em
    T\kern-.1667em\lower.7ex\hbox{E}\kern-.125emX}}

\begin{document}

\title{Feature Affinity Assisted Knowledge Distillation and Quantization of Deep Neural Networks on Label-Free Data }

\author{Zhijian Li, Biao Yang, Penghang Yin, Yingyong Qi, and Jack Xin 
\thanks{Zhijian Li, Biao Yang, Yingyong Qi, and Jack Xin are with Department of Mathematics, University of California, Irvine, CA, USA}
\thanks{Penghang Yin is with Department of Mathematics and Statistics, State University of New York at Albany, Albany, NY, USA}
\thanks{Corresponding author: Zhijian Li (e-mail: zhijil2@uci.edu)}
\thanks{This work was partly supported by NSF grants DMS-1924935, DMS-1952644, DMS-2151235, DMS-2208126; and a Qualcomm faculty award.}}

\markboth{}
{First A. Author \MakeLowercase{\textit{et al.}}: Bare Demo of IEEEtai.cls for IEEE Journals of IEEE Transactions on Artificial Intelligence}

\maketitle

\begin{abstract}
In this paper, we propose a  feature affinity (FA) assisted knowledge distillation (KD)  method to improve  quantization-aware training of deep neural networks (DNN).
The FA loss on intermediate feature maps of DNNs plays the role of teaching middle steps of a solution to a student instead of only giving final answers in the  conventional KD where the loss acts on the network logits at the output level. Combining logit loss and FA loss,  
we found that the quantized student network receives stronger supervision than from the labeled ground-truth data. The resulting FAQD is capable of compressing model on label-free data, which brings immediate practical benefits as pre-trained teacher models are readily available and  unlabeled data are abundant. In contrast, data labeling is often laborious and expensive. Finally, we propose a fast feature affinity (FFA) loss that accurately approximates FA loss with a lower order of computational complexity, which helps speed up training for high resolution image input.
\end{abstract}



\section{Introduction}
Quantization is one of the most popular methods for deep neural network compression, by projecting network weights and activation functions to lower precision thereby accelerate computation and reduce memory consumption.
However, there is inevitable loss of accuracy in the low bit regime. One way to mitigate such an issue is through knowledge distillation (KD \cite{hinton2015distilling}). In this paper, we 
study a feature affinity assisted 
KD so that the student and teacher networks not only try to match their logits at the output level 
but also 
match feature maps in the intermediate stages. 
This is similar to teaching a student through intermediate steps of a solution instead of just showing the final answer (as in  conventional KD \cite{hinton2015distilling}). Our method does not rely on ground truth labels while enhancing student network learning and closing the gaps between full and low precision models. 
\subsection{Weight Quantization of Neural Network}
 Quantization-aware training (QAT) searches the optimal model weight in training. Given an objective $L$, the classical QAT scheme (\cite{courbariaux2015binaryconnect, rastegari2016xnor}) is formulated as 
\begin{equation}
\begin{cases}
w^{t+1} = w^t - \nabla_u L (u^t),\\
u^{t+1} = \text{Quant($w^{t+1}$)},
\end{cases}
\label{eq:QAT}
\end{equation}
where Quant is projection to a low precision quantized space.
Yin et al. \cite{yin2018binaryrelax} proposed BinaryRelax, a relaxation form of QAT, which replaces the second update of (\ref{eq:QAT}) by
\begin{equation}
\begin{multlined}
u^{t+1} = \frac{w^{t+1}+\lambda^{t+1} \text{Quant($w^{t+1}$)}}{1+\lambda^{t+1}}, \\
\lambda^{t+1}=\eta \lambda^t \;\; \text{ with }\eta >1.
\end{multlined}
\label{eq:BR}
\end{equation}
Darkhorn et al. \cite{dockhorn2021demystifying} further improved (\ref{eq:BR}) by designing a more sophisticated learnable growing scheme for $\lambda^t$ and adding a learnable parameter into $\text{Quant($\cdot$)}$.
Polino et al. \cite{KDQ} proposed quantized distillation (QD), a QAT framework that leverages knowledge distillation for quantization. Under QD, the quantized model receives supervision from both ground truth (GT) labels and a trained teacher in float precision (FP). The objective function has the generalized form ($\alpha \in (0,1)$):
\begin{equation}\mathcal{L}_{QD} = \alpha \mathcal{L}_{KD} + (1-\alpha)\mathcal{L}_{GT}
\label{eq:QD}
\end{equation}
where $\mathcal{L}_{KD}$ is Kullback–Leibler divergence (KL) loss, and $\mathcal{L}_{GT}$ is negative log likelihood (NLL) loss. In order to compare different methods fairly, we introduce two technical terms: end-to-end quantization and fine-tuning quantization. End-to-end quantization is to train a quantized model from scratch, and fine-tuning quantization is to train a quantized model from a pre-trained float precision (FP) model. With the same method, the latter usually lands a better result than the former. Li et al.\cite{BRECQ} proposed a mixed quantization (a.k.a. BRECQ) that takes a pre-trained model and partially retrains the model on a small subset of data. 

\subsection{Activation Quantization}
\begin{figure}[ht!]
    \centering
    \includegraphics[scale=0.5]{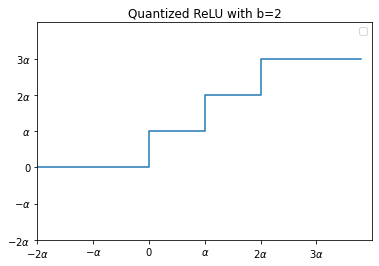}
    \caption{Plot of 2-bit quantized ReLU $\sigma(x, \alpha)$}
    \label{fig:qrelu}
\end{figure}
In addition to weight quantization, the inference of neural networks can be further accelerated through activation quantizaton. 
Given a resolution $\alpha>0$, a quantized ReLU activation function of bit-width $b\in\mathbb{N}$ is $
\sigma=\sigma(x, \alpha)$:
\begin{equation}
\sigma = 
\begin{cases}
 0 & x < 0\\
 k\alpha & (k-1)\alpha \leq x< k\alpha, \ \ 1\leq k \leq 2^b-1\\
 (2^b-1)\alpha & x \geq (2^b-1)\alpha
\end{cases}
\label{eq:qrelu}
\end{equation}
where the resolution parameter $\alpha$ is learned from data. A plot of $2$-bit quantized ReLU is shown in Fig. \ref{fig:qrelu}.
However, such quantized activation function leads to vanished gradient during training, which makes the standard backpropagation inapplicable. Indeed, it is clear that $\frac{\partial \sigma}{\partial x} = 0$ almost everywhere. Bengio et al. \cite{bengio2013estimating} proposed to use a straight through estimator (STE) in backward pass to handle the zero gradient issue. The idea is to simply replace the vanished $\frac{\partial \sigma}{\partial x}$ with a non-trivial derivative $\frac{\partial \tilde{\sigma}}{\partial x}$ of a surrogate function $\tilde{\sigma}(x,\alpha)$.  Theoretical studies on STE and convergence vs. recurrence issues of training algorithms have  been conducted in  (\cite{yin2019understanding,LYX23}). Among a variety of STE choices, a widely-used STE is the $x$-derivative of the so-called clipped ReLU \cite{cai2017deep} $\tilde{\alpha}(x,\alpha) = \min\{ \max\{x,0\}, (2^b-1)\alpha\}$, namely,
$$\frac{\partial \tilde{\sigma}}{\partial x}= \begin{cases}
  1 &   0 < x < (2^b-1)\alpha\\
  0 & \mbox{else}.
\end{cases}$$
In addition, a few proxies of $\frac{\partial \sigma}{\partial \alpha}$ have been proposed (\cite{choi2018pact, yin2019blended}). In this work, we follow \cite{yin2019blended} and use the three-valued proxy:
\begin{equation}
\frac{\partial \sigma}{\partial \alpha} \approx \begin{cases}
 0 & x\leq 0\\
 2^{b-1} & 0 < x< (2^b-1)\alpha\\
 2^b-1 & x \geq (2^b-1)\alpha.
\end{cases} 
\end{equation}
\subsection{Knowledge Distillation}
Several works have proposed to impose closeness of the probabilistic distributions between the teacher and student networks, 
e.g. similarity between feature maps.
A flow of solution procedure (FSP) matrix in \cite{fsp} measures the information exchange between two layers of a given model. Then $l_2$ loss regularizes the distance between FSP matrices of teacher and student in knowledge distillation. 
An attention transform (AT) loss \cite{atteionmap} directly measures the distance of feature maps outputted by teacher and student, which enhances the learning of student from teacher. Similarly, 
feature affinity (FA) loss \cite{FALoss}  measures the distance of two feature maps. 
In a dual learning framework for semantic segmentation \cite{FALoss}, the FA loss is applied on the output feature maps of a segmentation decoder and a high-resolution decoder. In \cite{FALossBiao}, FA loss is applied on multi-resolution paths in knowledge distillation of semantic segmentation models. It improves mean Average Precision of the lightweight student model. Given two feature maps with the same height and width (interpolate if different), $\mathbf{F}^S\in \mathbb{R}^{C_1\times H\times W}$  and $\mathbf{F}^T\in \mathbb{R}^{C_2\times H \times W}$, we first normalize the feature map along the channel dimension. Given a pixel of feature map $F_{i}\in \mathbb{R}^{C}$, we construct an affinity matrix $\mathbf{S}\in \mathbb{R}^{WH\times WH}$ as:
$$\mathbf{S}_{ij}=\|\vF_i- \vF_j\|_{\theta} :=\cos{\theta_{ij}}=\frac{\langle \vF_i, \vF_j\rangle}{||\vF_i||||\vF_j||}.
$$
where $\mathbf{\theta}_{ij}$ measures the angle between $F_i$ and $F_j$. Hence, the FA loss measures the similarity of
pairwise angular distance between pixels of two feature maps, which can be formulated as
\begin{equation}
L_{fa}(\mathbf{F}^S,\mathbf{F}^T)= \frac{1}{W^2H^2}||\mathbf{S}^T-\mathbf{S}^S||_2^2.
\label{eq:fa_loss}
\end{equation}
\subsection{Contributions}
In this paper, our main contributions are:
\medskip

\begin{enumerate}
    \item We find that using mean squares error (MSE) gives better performance than KL on QAT, which is a significant improvement of QD (\cite{KDQ}).
    \medskip
    
    \item We consistently improve the accuracies of various quantized student networks by imposing the FA loss on feature maps of each convolutional block. We also unveil the theoretical underpinning of  feature affinity loss in terms of the celebrated Johnson-Lindenstrass lemma for low-dimensional embeddings.
    \medskip
    
    \item We achieve state-of-art quantization accuracy on CIFAR-10, CIFAR-100, and Tiny ImageNet. Our FAQD framework {\it can train a quantized student network on unlabeled data} up to or exceeding the accuracy of its full precision counterpart. 
    \medskip
    
    \item We propose a randomized Fast FA (FFA) loss to accelerate the computation of training loss, and prove its convergence and error bound. 
\end{enumerate}
\subsection{Organization}
This paper is organized as follows: In Sec. II, we introduce the main objective of FAQD. In particular, we present feature affinity loss and go over the comparison between MSE and KL loss. In Sec. III, we numerically verify that FAQD outperforms baseline methods. In Sec. IV, we introduce Fast feature affinity loss and verify its acceleration to FAQD.
\section{Feature Affinity Assisted Distillation and Quantization}
\subsection{Feature Affinity Loss}
In quantization setting, it is unreasonable to require that $F^S$ be close to $F^T$, as they are typically in different spaces ($F^S \in \mathcal{Q}$ in full quantization) and of different dimensions. However, $F^S$ can be viewed as a compression of $F^T$ in dimension, and preserving information under such compression has been studied in compressed sensing. Researchers (\cite{ramasamy2015compressive, tremblay2016compressive}) have proposed to compress graph embedding to lower dimension  so that graph convolution can be computed efficiently. In K-means cluttering problem, several methods (\cite{becchetti2019oblivious, makarychev2019performance}) have been designed to project the data into a low-dimensional space such that
\begin{equation}
||\text{Proj}(\mathbf{x})-\text{Proj}(\mathbf{y})||\approx ||\mathbf{x}-\mathbf{y}||, \ \ \forall \; (\mathbf{x}, \mathbf{y}),
\label{eq:k_means}
\end{equation}
and so pairwise distances from data points to the centroids can be computed at a lower cost.
\medskip

In view of the feature maps of student model as a compression of teacher's feature maps, we impose a similar property in terms of pairwise angular distance:
$$
||\vF^S_i-\vF^S_j||_{\theta} \approx ||\vF^T_i-\vF^T_j||_{\theta}, \; \forall \, (i, j)
$$
which is realized by minimizing the feature affinity loss. 
On the other hand, a Johnson–Lindenstrauss (JL \cite{johnson1986extensions}) like lemma can guarantee that we have student's feature affinity matrix close to the teacher's, provided that the number of channels of student network is not too small. In contrast, the classical JL lemma states that a set of points in a high-dimensional space can be embedded into a space of much lower dimension in such a way that the \emph{Euclidean} distances between the points are nearly preserved. To tailor it to our application, we prove the following JL-like lemma in the angular distance case: 

\begin{theorem}[Johnson–Lindenstrauss lemma, Angular Case]
\label{JD}
Given any $\epsilon \in (0,1)$, an embedding matrix $\mathbf{F}\in \mathbb{R}^{n\times d}$, for $k \in (16\epsilon^{-2}\ln{n}, d)$, there exists a linear map $T(\mathbf{F})\in \mathbb{R}^{n\times k }$ so that
\begin{equation}
\begin{multlined}
(1-\epsilon)||\mathbf{F}_i-\mathbf{F}_j||_{\theta}\leq ||T(\mathbf{F})_i-T(\mathbf{F})_j||_{\theta}\\
\leq (1+\epsilon)||\mathbf{F}_i-\mathbf{F}_j||_{\theta}, \ \ \forall \ \ 1\leq i,j \leq n
\end{multlined}
\end{equation}
where $||\vF_i-\vF_j||_{\theta} = \frac{\langle \vF_i,\vF_j \rangle }{\|\vF_i\|\|\vF_j\|}$ is the angular distance.
\end{theorem}
\medskip

It is thus possible to reduce the embedding dimension down from $d$ to $k$, while roughly preserving the pairwise angular distances between the points. In a convolutional neural network, we can view intermediate feature maps as $\mathbf{F}^{S} \in \mathbb{R}^{HW\times C_1}$ and $\mathbf{F}^{T}\in \mathbb{R}^{HW\times C_2}$, and feature affinity loss will help the student learn a compressed feature embedding. The  
FA loss can be flexibly placed between teacher and student in different positions (encoder/decoder, residual block, etc.) for different models. In standard implementation of ResNet, residual blocks with the same number of output channels are grouped into a sequential layer. We apply FA loss to the features of such layers.
$$
\mathcal{L}_{FA}=\sum_{l=1}^L L_{fa}(\bold{F}_l^T, \bold{F}_l^S)
$$
where $\bold{F}_l^T$ and $\bold{F}_l^S$ are the feature maps of teacher and student respectively. For example, the residual network family of ResNet20, ResNet56, ResNet110, and ResNet164 have $L=3$, whereas the family of ResNet18, ResNet34, and ResNet50 have $L=4$.

\subsection{Choice of Loss Functions}

In this work, we propose two sets of loss function choices for the end-to-end quantization and pretrained quantization, where end-to-end quantization refers to having an untrained student model with randomly initialized weights. 
We investigate both scenarios of quantization and propose two different strategies for each.
\medskip

The Kullback–Leibler divergence (KL) is a metric of the similarity between two probabilistic distributions. Given a ground-truth distribution $P$, it computes the relative entropy of a given distribution $Q$ from $P$:
\begin{equation}
    \mathcal{L}_{KL}(P||Q) = \sum_{x\in \mathcal{X}}P(x)\ln{\frac{P(x)}{Q(x)}}.
    \label{eq:KL}
\end{equation}
While KD is usually coupled with KL loss (\cite{KDQ, hinton2015distilling}), it is not unconventional to choose other loss functions. Kim et al. \cite{kim2021comparing} showed that MSE, in certain cases, can outperform KL in the classic teacher-student knowledge distillation setting. KL loss is also widely used for trade-off between accuracy and robustness under adversarial attacks, which can be considered as self-knowledge distillation. Given a classifier $f$, an original data point $\bf{x}$ and its adversarial example $\bf{x'}$, TRADES \cite{zhang2019theoretically} is formulated as
$$L_{TRADES}=\mathcal{L}_{CE}(f(\bf{x}), y)+\mathcal{L}_{KL}(f(\bf{x})|| f(\bf{x'})).$$
Li et al. \cite{li2021integrated} showed that $L_{CE}\big(f(\mathbf{x}'), y\big)$ outperforms $\mathcal{L}_{KL}(f(\bf{x})|| f(\bf{x'}))$ both experimentally and theoretically.
\medskip

Inspired by the studies above, we conduct experiments on different choices of the loss function. We compare KD on quatization from scratch (end-to-end). As shown in Tab. \ref{tab:KL_vs_MSE}, MSE outperforms KL in quantization.
\begin{table}[ht!]
    \centering
    \begin{tabular}{c|c|c|c|c}
        Student & Teacher & 1-bit & 2-bit & 4-bit\\
        \hline
        \hline
         \multicolumn{5}{c}{$\mathcal{L}_{KD}=$ KL loss in (\ref{eq:QD})} \\
         \hline
         ResNet20&ResNet110 & 89.06\% & 90.86\%&92.01\%\\
         \hline
         \multicolumn{5}{c}{$\mathcal{L}_{KD}=$ MSE in (\ref{eq:QD})} \\
         \hline
         ResNet20&ResNet110 & \bf{90.00}\% & \bf{91.01}\%&\bf{92.17}\%\\
         \hline
    \end{tabular}
    \caption{Comparision of KL loss and MSE loss on CIFAR-10 data set. All teachers are pre-trained FP models, and all students are initial models (end-to-end quantization).}
    \label{tab:KL_vs_MSE}
\end{table}

On the other hand, we find that KL loss works better for fine-tuning quantization. One possible explanation is that when training from scratch, the term $\ln{\frac{P(x)}{Q(x)}}$ is large. However, the derivative of logarithm is small at large values, which makes it converge slower and potentially worse. On the other hand, when $\frac{P(x)}{Q(x)}$ is close to 1, the logarithm has sharp slope and converges fast. 

\subsection{Feature Affinity Assisted Distillation and Quantization}
\begin{figure*}
    \centering
    \includegraphics[width = 15cm, height = 12cm]{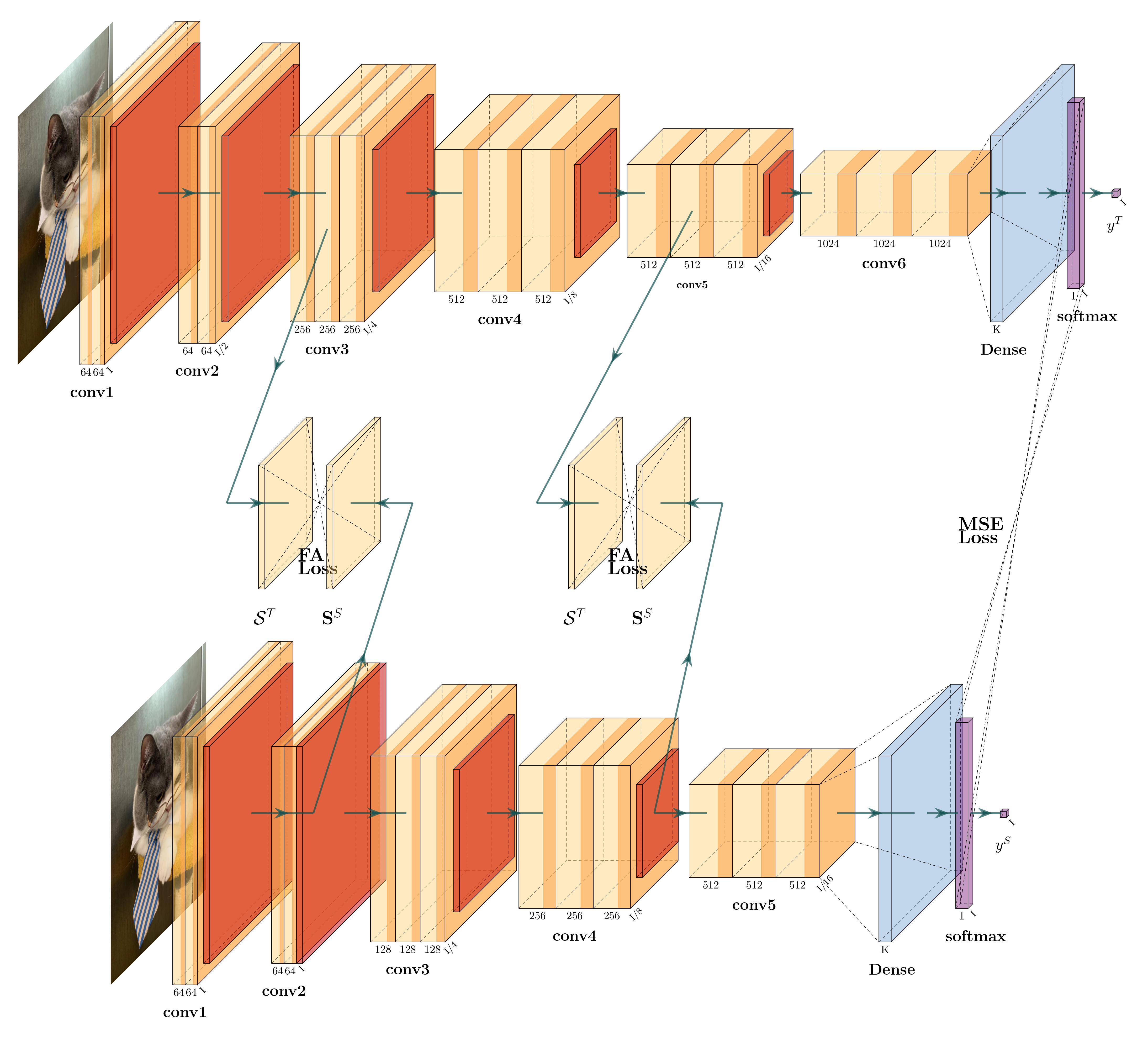}
    \caption{FAQD framework. The intermediate feature maps are supervised by FA loss, and the raw logits by MSE loss.}
    \label{fig:Illustration of KD }
\end{figure*}
Inspired by previous studies (\cite{KDQ, QKD, BRECQ}), we propose a feature affinity assisted quantized distillation (FAQD). The end-to-end quantization objective function is formulated as:
\begin{equation}
\begin{multlined}
\mathcal{L} = \alpha \, \mathcal{L}_{KD}+\beta\, \mathcal{L}_{FA}+\gamma \, \mathcal{L}_{GT}\\
=\alpha\, \mathcal{L}_{MSE}\big(f^T(\mathbf{x}), f^S(\mathbf{x})\big)+\beta\, \sum_{l=1}^{L} \mathcal{L}_{fa}(\mathbf{F}_l^T, \mathbf{F}_l^S)\\
+\gamma \, \mathcal{L}_{NLL}(f^S(\mathbf{x}), y).
\end{multlined}
\label{eq: FAQD1}
\end{equation}
In fine-tuning quantization, we replace MSE loss in \eqref{eq: FAQD1} by KL divergence loss. In FAQD, the student model learns not only the final logits of the teacher but also the intermediate extracted feature maps of the teacher using feature affinity norm computed as in \cite{FALoss}.

In addition to (\ref{eq: FAQD1}), we also propose a label-free objective which does not require the knowledge of labels:
\begin{equation}
\mathcal{L}_{\text{label-free}}=\alpha \mathcal{L}_{MSE}\big(f^T(\mathbf{x}), f^S(\mathbf{x})\big)+\beta\sum_{l=1}^{L} \mathcal{L}_{fa}(\mathbf{F}_l^T, \mathbf{F}_l^S).
\label{eq: FAQD2}
\end{equation}
Despite the pre-trained computer vision models being available from cloud service such as AWS and image/video data abundantly collected, the data labeling is still expensive and time consuming. Therefore, a label-free quantization framework has significant value in the real world. 
In this work, we verify that the FA loss can significantly improve KD performance.
The label-free loss in Eq. \eqref{eq: FAQD2} can outperform the baseline methods in Tab. \ref{tab:baseline1}  as well as the prior  supervised QD in \eqref{eq:QD}. 
\section{Experimental Results}
\begin{table}[ht!]
    \centering
\begin{tabular}{c|c|c|c}
              Method &1-bit&2-bit&4-bit   \\
    \hline 
    \hline
    \multicolumn{4}{c}{Model: ResNet20}\\
    \hline
       QAT (\cite{courbariaux2015binaryconnect, rastegari2016xnor}) & 87.07\%&90.26\%&91.47\%\\
     \hline
       BinaryRelax  \cite{yin2018binaryrelax} &88.64\%&90.47\%&91.75\%\\
     \hline
       QD \cite{KDQ} &89.06\%&90.86\%&92.01\%\\
     \hline
        DSQ \cite{gong2019differentiable}&90.24\%&91.06\%&91.92\%\\
      \hline
        BRECQ$^*$ \cite{BRECQ} & N/A &88.10\%&89.01\%\\
      \hline
     
\end{tabular}
\caption{End-to-end quantization accuracies of some existing quantization-aware training methods on CIFAR-10 dataset. To stick with the original work, we apply channel-wise quantization in BRECQ, denoted by $*$. All the other methods are under layer-wise quantization.}
\label{tab:baseline1}
\end{table}
In Tab. \ref{tab:baseline1}, we listed the performance of previous methods mentioned in the introduction section. We would like to remark that the BRECQ results are from channel-wise quantization. Namely, each channel of a convolutional layer has its own float scaler and projection map. All other results in Tab. \ref{tab:baseline1} are layer-wise quantization.\\
All experiments reported here were conducted on a desktop with Nvidia RTX6000
8GB GPU card at UC Irvine. 
\subsection{Weight Quantization}
In this section we test FAQD on the dataset CIFAR-10. First, we experiment on fine-tuning quantization. The float precision (FP) ResNet110 teaches ResNet20 and ResNet56. The teacher has 93.91\% accuracy, and the two pre-trained models have accuracy 92.11\% and 93.31\% respectively. While both SGD and Adam optimization work well on the problem, we found KL loss with Adam slightly outperform SGD in this scenario. The objective is
$$\mathcal{L} = \mathcal{L}_{KL}+\mathcal{L}_{FA}$$
for the label-free quantization. When calibrating the ground-truth label, the cross-entropy loss $\mathcal{L}_{NLL}$ is used as the supervision criterion. 

\begin{table}[ht!]
    \centering
\begin{tabular}{c|c|c|c}
    \hline
    \hline
    \multicolumn{4}{c}{\bf{Cifar-10}}\\
    \hline 
     \multicolumn{4}{c}{Teacher ResNet110: 93.91\%}\\
     \hline
    \multicolumn{4}{c}{Pre-trained FP ResNet20: 92.21\%}\\
    \hline
    Method & 1-bit&2-bit&4-bit   \\
     \hline
   Label-free FAQD& 89.97\%&91.40\%&92.55\%\\
     \hline
   FAQD with Supervision&  90.92\%&91.93\%&92.74\%\\
     \hline
     \multicolumn{4}{c}{\bf{Cifar-100}}\\
    \multicolumn{4}{c}{Teacher ResNet164: 74.50\%}\\
    \multicolumn{4}{c}{Pre-trained FP ResNet110:72.96\%}\\
    \hline
     Method &1-bit&2-bit&4-bit   \\
    \hline
    label-free FAQD &73.33\%&75.02\%&75.78\%\\
    \hline
     FAQD with Supervision &73.35\%&75.24\%&76.10\%\\
     \hline
     \multicolumn{4}{c}{\bf{Tiny ImageNet}}\\
    \multicolumn{4}{c}{Teacher ResNet34: 65.60\%}\\
    \multicolumn{4}{c}{Pre-trained FP ResNet18: 64.23\% }\\
    \hline
     Method &1-bit&2-bit&4-bit   \\
    \hline
    label-free FAQD &64.89\%&65.02\%&65.69\%\\
    \hline
     FAQD with Supervision &65.77\%&66.49\%&66.62\%\\
     \hline
\end{tabular}

\caption{Fine-tuning knowledge distillation for quantization of all convolutional layers. 
}
\label{tab:pretrained}
\end{table}
For end-to-end quantization, we found that MSE loss performs better than KL loss. Adam optimization struggles to reach acceptable performance on end-to-end quantization (with either KL or MSE loss). 
We further test the performance of FAQD on larger dataset CIFAR-100 where an FP ResNet 164 teaches a quantized ResNet110. We report the accuracies for both label-free and label-present supervision. 
We evaluate FAQD on both fine-tuning quantization and end-to-end quantization.
\begin{table}[ht!]
    \centering
\begin{tabular}{c|c|c|c}
    \hline 
    \hline
    \multicolumn{4}{c}{\bf{Cifar-10}}\\
    \multicolumn{4}{c}{Teacher ResNet110: 93.91\%}\\
        \hline
     Method &1-bit&2-bit&4-bit   \\
    \hline 
    \multicolumn{4}{c}{FP student ResNet20: 92.21 \%} \\
     \hline
     Label-free FAQD &89.88\%&91.23\%&92.19\%\\
     \hline
     FAQD with Supervision &90.56\%&91.65\%&92.43\%\\
     \hline
    \multicolumn{4}{c}{\bf{Cifar-100}}\\
    \multicolumn{4}{c}{Teacher ResNet164: 74.50\%}\\
    \hline
             Method &1-bit&2-bit&4-bit   \\
    \hline
    label-free FAQD &72.78\%&74.35\%&74.90\%\\
    \hline
     FAQD with Supervision&73.35\%&74.40\%&75.31\%\\
     \hline
     \multicolumn{4}{c}{\bf{Tiny ImageNet}}\\
    \multicolumn{4}{c}{Teacher ResNet34: 65.60\%}\\
    \hline
     Method &1-bit&2-bit&4-bit   \\
    \hline
    label-free FAQD &64.37\%&65.05\%&65.40\%\\
    \hline
     FAQD with Supervision &65.13\%&65.67\%&65.92\%\\
     \hline
\end{tabular}
\caption{End-to-end FAQD of ResNet110 on CIFAR-100. The accuracy of 4-bit label-free quantization surpasses 72.96\% of FP ResNet110 and is close to FP ResNet164.}
\label{tab:end-to-end}
\end{table}
In the CIFAR-100 experiment, the teacher ResNet164 has 74.50\% testing accuracy. For the pretrained FAQD, the FP student ResNet110 has 72.96\% accuracy. As shown in Tab. \ref{tab:pretrained} and Tab. \ref{tab:end-to-end}, FAQD has surprisingly superior performance on CIFAR-100. The binarized student almost reaches the accuracy of FP model, and the 4-bit model surpasses the FP teacher.

\subsection{Full Quantization}
\begin{table}[ht!]
    \centering
\begin{tabular}{c|c|c}
              Method &1W4A&4W4A   \\
    \hline 
    \hline
    \multicolumn{3}{c}{Model: ResNet20}\\
     \hline
       BinaryRelax  \cite{yin2018binaryrelax} &89.22\%&91.37\%\\
     \hline
       QD \cite{KDQ} &90.15\%&92.06\%\\
     \hline
        BCGD \cite{yin2019blended}&89.98\%&91.65\%\\
      \hline
        BRECQ$^*$ \cite{BRECQ} & N/A &88.71\%\\
      \hline
     
\end{tabular}
\caption{Fine-tuning full quantization results of existing methods on CIFAR-10. The $*$ means the same as in Tab. 2.}
\label{tab:baseline2}
\end{table}
In this section, we extend our results to full quantization where the activation function is also quantized. In Tab. \ref{tab:baseline2}, we list the fine-tuning results from aforementioned methods. Among the methods in Tab. \ref{tab:baseline2}, only Quantized Distillation (QD) is stable under end-to-end full quantization. We extend our results to the tiny Tiny ImageNet dataset, which contains 100K downsampled 64$\times$64 images across 200 classes for training. To simulate ImageNet, we interpolate the resolution back to the original 224$\times$224. As shown in Tab. \ref{tab:fullq}, the 4W4A fune-tuning quantization has accuracy similar to float ResNet20. Meanwhile, we close the long existing performance gap \cite{DSQ} when reducing activation precision to 1-bit, as the accuracy drop is linear (with respect to activation precision) and small. When fine-tuning a fully quantized model, we follow a two-step process. First, we train an activation quantized model with floating-point weights. Subsequently, we apply full quantization using the FAQD. This technique proves to be essential, especially when scaling up the Tiny ImageNet dataset. In our experiments, we observed  the following phenomenon when replacing all ReLU activation functions with 1-bit Quantized ReLU. For a pretrained 32A32W ResNet20 model, originally trained on CIFAR-10, the accuracy dropped to 80.03\% from its original accuracy of 92.21\%. However, when working with a pretrained ResNet-18 model on the Tiny ImageNet dataset, the accuracy plummeted to 0.62\% from its initial accuracy of 64.23\%.
\begin{table}[ht!]
    \centering
    \begin{tabular}{c|c|c|c}
    \hline
    \hline
    \multicolumn{4}{c}{\bf{CIFAR-10}}\\
    \multicolumn{4}{c}{Pretrained ResNet20: 1A32W-91.89\%, 4A32W-92.01\%}\\
    \hline
    Model & pre-trained &1W1A  & 4W4A \\
         
         \hline
         ResNet20 & No &N/A&91.07\%\\
         ResNet20 & Yes &89.70\%& 92.53\%\\
         \hline
    \multicolumn{4}{c}{\bf{CIFAR-100}}\\
    \multicolumn{4}{c}{Pretrained ResNet56: 1A32W-70.96\%, 4A32W-71.42\%}\\
        \hline
        Model & pre-trained &1W1A  & 4W4A \\ 
         \hline
        ResNet56&No&N/A&68.84\%\\
        ResNet56&Yes&68.18&73.53\%\\
    \hline
    \multicolumn{4}{c}{\bf{Tiny ImageNet}}\\
    \multicolumn{4}{c}{Pretrained ResNet18: 1A32W-63.82\%, 4A32W-64.15\%}\\
        \hline
        Model & pre-trained &1W1A  & 4W4A \\ 
         \hline
        ResNet18&No&N/A&64.67\%\\
        ResNet18&Yes&65.01&65.55\%\\
    \hline
    \end{tabular}
    \\
    \caption{End-to-end and fine-tuning full quantization on CIFAR-10, CIFAR-100 and Tiny ImageNet, with teacher networks same as in Tab. 4.}
    \label{tab:fullq}
\end{table}

\section{Fast Feature Affinity Loss}
\subsection{Proposed Method}
\begin{figure}[ht!]
    \centering
    \includegraphics[scale =0.6]{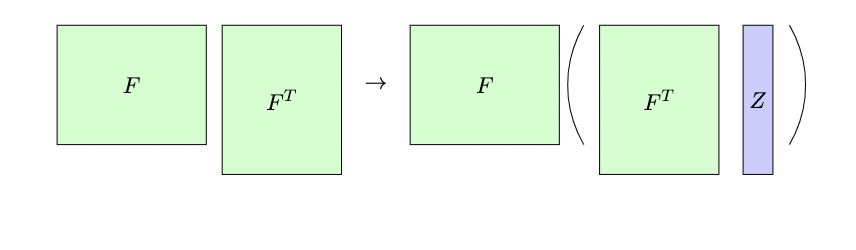}
    \caption{Fast feature affinity loss with a low-rank random matrix $Z$.}
    \label{fig:FFA}
\end{figure}
Despite the significant increase of KD performance, we note that introducing FA loss will increase the training time. If we normalize the feature maps by row beforehand, computing FA loss between multiple intermediate feature maps can be expensive.\\
\begin{equation}
\label{eq:n_fa_1}
\cL_{fa}(F_1,F_2) = \|F_1F_1^T - F_2F_2^T\|_2^2.
\end{equation}
As we freeze the pre-trained teacher, feature map of the teacher model $F_1 = f^{T}(\vx)$ is a constant, in contrast to student feature map $F_2 = f^S(\Theta, \vx)$. Denote $\mathbf{S}_1 = F_1F_1^T \in \mathbb{R}^{WH\times WH}$ and $g(\Theta, \vx)=f^S(\Theta, \vx)[f^S(\Theta, \vx)]^T$. The feature affinity can be formulated as 
\begin{equation}
    \cL_{fa}(\Theta) = \frac{1}{|\mathcal{X}|}\sum_{\vx \in \mathcal{X}}\|\mathbf{S}_1 - g(\Theta, \vx)\|_2^2.
    \label{eq:fa}
\end{equation}
Computing $\mathbf{S}_1$ and $g(\Theta, X)$ requires $\mathcal{O}(W^2H^2C)$ complexity each (C is the number of channels), which is quite expensive. We introduce a random estimator of $L_{ffa}(\Theta)$:
\begin{equation}
    \cL_{ffa}(F_1, F_2, \mathbf{z}) = \frac{1}{|\mathcal{X}|}\sum_{\vx \in \mathcal{X}}\|(\mathbf{S}_1 - g(\Theta, \vx))\mathbf{z}\|_2^2,
    \label{eq:ffa_one}
\end{equation}
where $\mathbf{z}\in\mathbb{R}^{HW}$ is a vector with i.i.d unit normal components $\cN(0,1)$. We show below that Eq. (\ref{eq:ffa_one}) is an unbiased estimator of FA loss (\ref{eq:fa}). 
\begin{prop}
$$\mathbb{E}_{\mathbf{z}\sim \cN(0,1)} [\mathcal{L}_{ffa}(F_1, F_2, \mathbf{z})] = \mathcal{L}_{fa}(\Theta).
$$
\end{prop}
This estimator can achieve computing complexity $\mathcal{O}(HWC)$ by performing two matrix-vector multiplication  
$F_1\big(F_1^T\mathbf{z}\big)$.
We define the Fast Feature Affinity (FFA) loss to be the $k$ ensemble of (\ref{eq:ffa_one}):
\begin{equation}
    \cL_{ffa, k}(\Theta) = \frac{1}{|\mathcal{X}|}\sum_{\vx \in \mathcal{X}}\frac{1}{k}\|(\mathbf{S}_1 - g(\Theta, \vx ))Z_k\|_2^2
    \label{eq:ffa}
\end{equation}
where $Z_k\in \mathbb{R}^{HW\times k}$ with i.i.d $\mathcal{N}(0,1)$ components, and we have $k\ll WH$. The computational complexity of $\mathcal{L}_{ffa, k}(\Theta)$ is $\mathcal{O}(kWHC)$.
\medskip

Finally, we remark that FFA loss can accelerate computation of pairwise Euclidean distance in dimensional reduction such as in \eqref{eq:k_means}. The popular way to compute the pairwise distance of rows for a matrix $A\in\mathbb{R}^{n\times c}$ is to broadcast the vector of row norms and compute $AA^T$. Given the row norm vector $v=(\|A_1\|^2, \cdots, \|A_n\|^2)$, 
the similarity matrix $(\mathbf{S}_{ij})$,
$\mathbf{S}_{ij} = \|A_i-A_j\|^2$, is computed as 
$$\mathbf{S} = \mathbf{1}\otimes v -2AA^T + v \otimes \mathbf{1}.$$ The term $2AA^T$ can be efficiently approximated by FFA loss.
\subsection{Experimental Results}
\begin{figure}[ht!]
\begin{subfigure}{0.5\textwidth}
\includegraphics[scale =0.15]{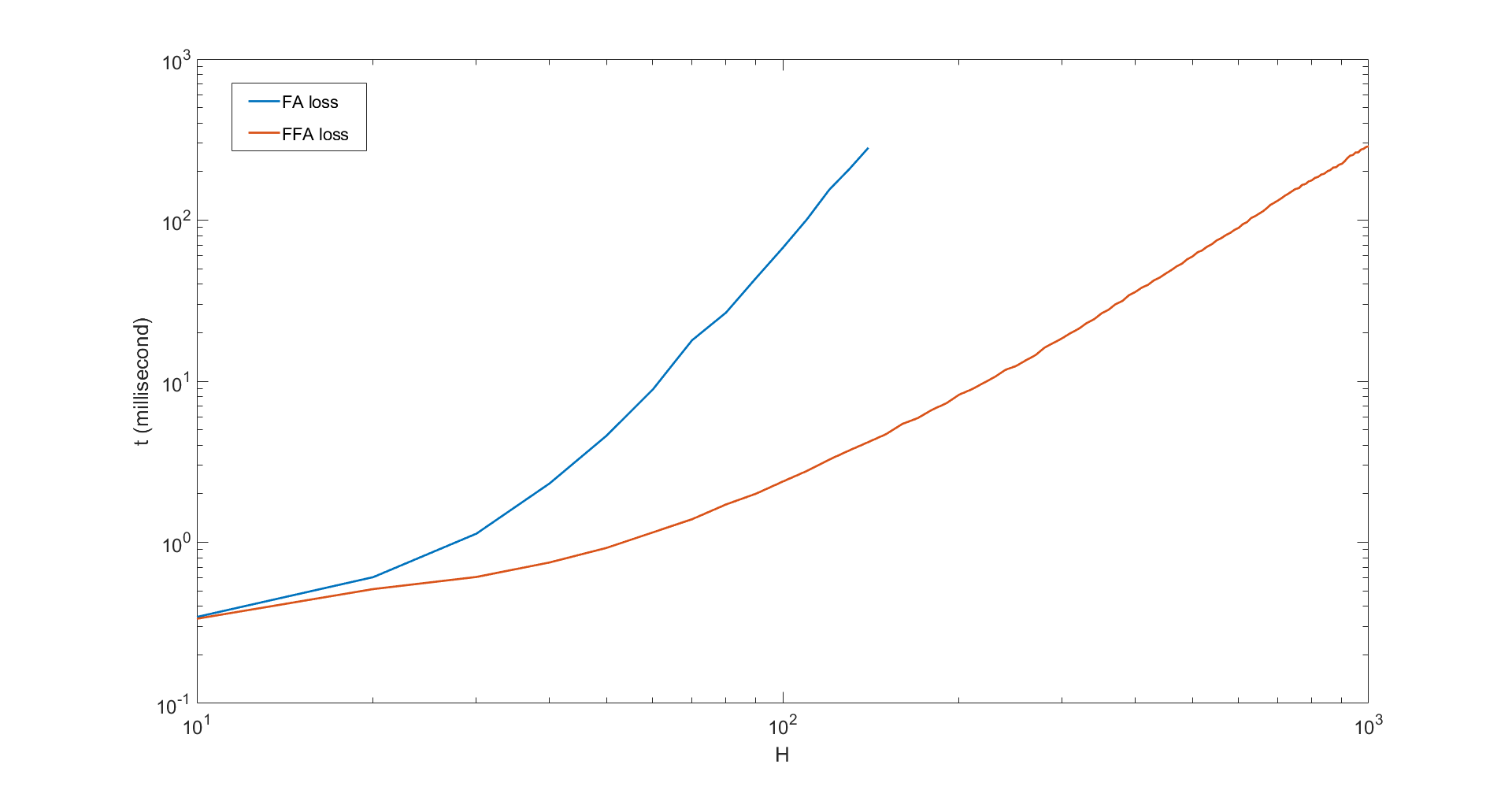}
\caption{Log-log plot for inference time of FA loss and FFA loss. }
\label{fig:loglog}  
\end{subfigure}
\hfill
\begin{subfigure}{0.5\textwidth}
\includegraphics[scale =0.15]{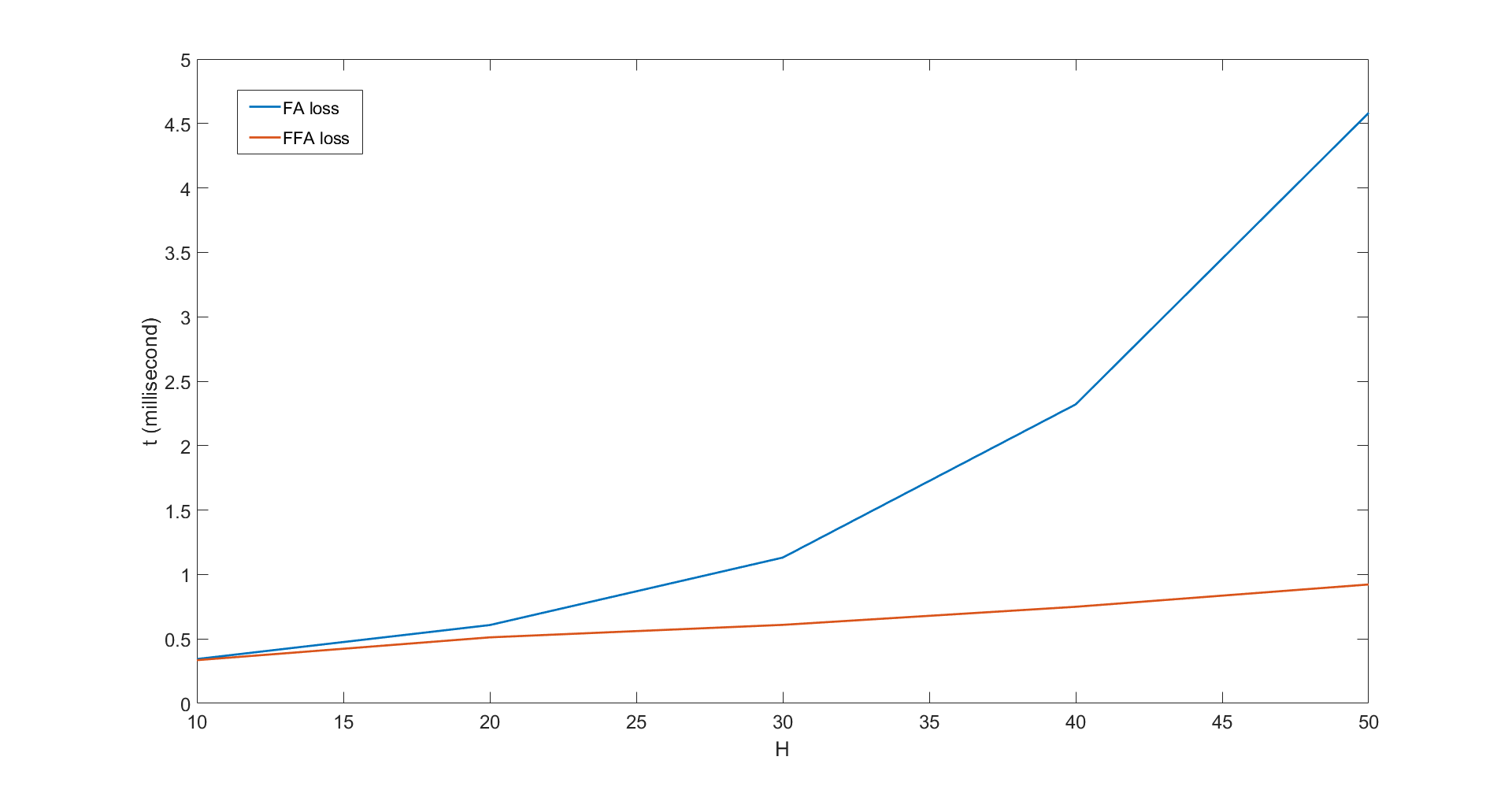}
\caption{Zoomed in plot for inference time of FA loss and FFA loss. }
\label{fig:fa_zoom}  
\end{subfigure}
\caption{Plots for inference time of FA loss and FFA loss with $k=1$.}
\label{fig:faplot}
\end{figure}

We test Fast FA loss on CIFAR-10 and Tiny ImageNet. As mentioned in the previous section, ResNet-20 has 3 residual blocks. The corresponding width and height for feature maps are 32, 16, and 8, $H=W$ for all groups, so the dimension ($HW$) of similarity matrices are 1024, 256, and 64. We test the fast FA loss with the number of ensemble $k=$ 1, 5, and 15. The results are shown in Tab. \ref{tab:ffa}. Meanwhile, FFA has added training time for each step. When $k=1$, the accuracies are inconsistent due to large variance. With too few samples in the estimator, the fast FA norm is too noisy and jeopardizes distillation. At $k=5$, the fast FA loss stabilizes and the accuracy improves towards that of the baseline, $\mathcal{L}=\mathcal{L}_{MSE}+\mathcal{L}_{CE}$ in Tab. \ref{tab:KL_vs_MSE}. When $k$ increases to $15$, the performance of fast FA loss is comparable to that of the exact FA loss. 
Moreover, we experiment with the time consumption for computing FA loss and FFA loss. We plot the time in log scale vs. $H$, ($H=W$) for feature maps.  Theoretical time complexity for computing exact FA loss is $\mathcal{O}(H^4)$ and that for FFA loss is  $\mathcal{O}(H^2)$. Fig. \ref{fig:faplot}(a) shows the agreement with the theoretical estimate. 
\begin{table}[ht!]
    \centering
    \begin{tabular}{c|c|c}
    \hline
    \hline
    Dataset& CIFAR-10& Tiny ImageNet\\
    \hline
    Model & ResNet20  & ResNet18 \\
    \hline
    \hline
    Ensemble Number $k$ &\multicolumn{2}{c}{Fast FA Loss Accuracy}\\
         
         \hline
          1 (ResNet20)/1 (ResNet18) &88.89$\pm$2.95\% & 52.32$\pm$ 4.35\%\\
         5/40 &90.55\%  & 56.12\%\\
         15/80 &90.72\% & 61.12\%\\
         \hline
    Ensemble Number $k$&\multicolumn{2}{c}{Fast FA Loss Training Time Per Epoch}\\
    \hline
    1/1 &29.71s &5m19s\\
          5/40 & 29.77s &  5m32s\\
          15/80 &30.74s &  5m51s\\
    \hline
     Ensemble Number $k$ &\multicolumn{2}{c}{Exact FA Loss Training Time Per Step}\\
     \hline
     N/A& 36.17s & 7m36s\\
     \hline  
    \end{tabular}
    \caption{4A4W FFA accuracy and training time per epoch for ResNet20 on CIFAR-10 and ResNet18 on Tiny ImageNet, with teacher networks same as in Tab. 4. The FFA loss accelerates training and approaches the performance of exact FA loss with a proper choice of the ensemble number $k$.}
    \label{tab:ffa}
\end{table}
The larger the $H$, the more advantageous the FFA loss. For (medical) images with resolutions in the thousands, the FFA loss will have significant computational savings. In Tab. \ref{tab:ffa}, we report training time per epoch. We train models 200 epochs with cosine annealing learning rate.

\subsection{Theoretical Analysis of FFA Loss}
As shown in Proposition 4.1, the FFA loss is a $k$-ensemble unbiased estimator of FA loss. By the strong law of large numbers, the FFA loss converges to the exact FA loss with probability 1.
\begin{theorem}
For given $\Theta$, suppose that $|\cL_{fa}(\Theta)|< \infty$, then
$$\forall \epsilon >0, \exists N \ \ s.t. \ \ \forall k>N, \ \ |\cL_{ffa, k}(\Theta)-\cL_{fa}(\Theta)|<\epsilon.$$ Namely, the FFA loss converges to FA loss pointwise:
$$\forall \, \Theta, \; \lim_{k\to \infty}\cL_{ffa, k}(\Theta) = \cL_{fa}(\Theta).$$  
\end{theorem}
We also establish the following error bound for finite $k$.
\begin{prop}
    $$\mathbb{P}\big(|\cL_{ffa, k}(\Theta)-\cL_{fa}(\Theta)|>\epsilon\big)\leq \frac{C}{\epsilon^2k},$$
    where $C\leq 3\, \|\cL_{fa}(\Theta)\|_2^4$.
\end{prop}
\medskip

Proposition 4.2 says that the probability that the FFA estimation has an error beyond a target value decays like $\mathcal{O}(\frac{1}{k})$.
The analysis guarantees the accuracy of FFA loss as an efficient estimator of FA loss. Another question one might ask is whether minimizing the FFA loss is equivalent to minimizing the FA loss. Denote $\Theta^*=\argmin L_{fa}(\Theta)$ and $\Theta_k^*=\argmin L_{ffa, k}(\Theta)$, and assume the minimum is unique for each function. In order to substitute FA loss by FFA loss, one would hope that $\Theta_k^*$ converges to $\Theta^*$. Unfortunately, the point-wise convergence in Theorem 4.1 is not sufficient to guarantee the convergence of the optimal points, as a counter-example can be easily constructed. In the rest of this section, we show that such convergence can be established under an additional assumption. 
\medskip

\begin{theorem}[Convergence in the general case]
Suppose that $\cL_{ffa, k}(\Theta)$ converges to $\cL_{fa}(\Theta)$ uniformly, that is
$$\forall \, \epsilon>0, \exists\, N \ \ s.t. \ \ \forall \, k>N,\ \
|\cL_{ffa, k}(\Theta)-\cL_{fa}(\Theta)|<\epsilon$$
and $|\cL_{fa}(\Theta)|< \infty, \; \forall \Theta.$
Then
\begin{equation}
    \lim_{k \to \infty}||\Theta_k^*-\Theta^*||^2 = 0. 
\end{equation}
\end{theorem}
The uniform convergence assumption can be relaxed if $\mathcal{L}_{fa}$ is convex in $\Theta$. A consequence of Theorem 4.2 is below.
\medskip

\begin{corollary}[Convergence in the convex case]
 Let $L_{fa}:\mathbb{R}^n\to \mathbb{R}$ be convex and $L$-smooth, and that  $\exists\,$ constant $M>0$ such that $||\Theta_k^*||\leq M,\, \forall k$. Then  $\mathcal{L}_{ffa, k} \text{ is also convex}$ for any $k$,
 and $\lim_{k\rightarrow \infty} ||\Theta_k^* - \Theta^*||^2=0.$ 
\end{corollary}
\section{Conclusion}
We presented FAQD, a feature assisted (FA) knowledge distillation method for quantization-aware training. It couples MSE loss with FA loss and significantly improves the accuracy of the quantized student network. FAQD applies to both weight only and full quantization, and outperforms baseline Resnets on CIFAR-10/100 and Tiny ImageNet.
We also analyzed an efficient randomized approximation (FFA) to the FA loss for large dimensional feature maps, which provided theoretical foundation for FFA loss to benefit future model training on high resolution images in applications.

\section{Appendix}
\textbf{Proof of Theorem 2.1:} 
It suffices to prove that for any set of $n$ unit vectors in $\mathbb{R}^d$, there is a linear map nearly preserving pairwise angular distances, because the angular distance is scale-invariant.

Let $T$ be a linear transformation induced by a random Gaussian matrix $\frac{1}{\sqrt{k}}A\in \mathbb{R}^{k\times d}$ such that $T(\vF) = \vF A^T$. 
Define the events $\mathcal{A}_{ij}^- =\{T: (1-\epsilon)\|\vF_i-\vF_j\|^2\leq \|T(\vF)_i-T(\vF)_j\|^2\leq (1+\epsilon)\|\vF_i-\vF_j\|^2 \text{ fails}\}$ and $\mathcal{A}^+_{ij} =\{T: (1-\epsilon)\|\vF_i+\vF_j\|^2\leq \|T(\vF)_i+T(\vF)_j\|^2\leq (1+\epsilon)\|\vF_i+\vF_j\|^2 \text{ fails}\}$. 

Following the proof of the classical JL lemma in the Euclidean case \cite{vempala2005random}, we have:
\begin{equation}
 P(\mathcal{A}^-_{ij})\leq   2e^{-\frac{(\epsilon^2-\epsilon^3)k}{4}},\ \    P(\mathcal{A}^+_{ij})\leq   2e^{-\frac{(\epsilon^2-\epsilon^3)k}{4}}.
\end{equation}
Let $\mathcal{B}_{ij}=\{T: |\vF_i\cdot \vF_j -  T(\vF)_i\cdot T(\vF)_j |>\epsilon\}$, where $\cdot$ is the shorthand for inner product.
We show that $\mathcal{B}_{ij}\subset A^-_{ij}\cup \mathcal{A}^+_{ij}$ for $\|F_i\|=\|F_j\|=1$ by showing ${\mathcal{A}^-_{ij}}^C\cap {\mathcal{A}^+_{ij}}^C\subset \mathcal{B}_{ij}^C$. 

If ${\mathcal{A}^-_{ij}}^C\cap {\mathcal{A}^+_{ij}}^C$ holds, we have
\begin{align*}
& 4T(\vF)_i\cdot T(\vF)_j \\
= & \|T(\vF)_i+T(\vF)_j\|^2-\|T(\vF)_i-T(\vF)_j\|^2 \\
\leq & (1+\epsilon)\|\vF_i+\vF_j\|^2-(1-\epsilon)\|\vF_i-\vF_j\|^2 \\
= & 4\vF_i \cdot \vF_j + 2\epsilon(\|\vF_i\|^2+\|\vF_j\|^2) \\
= & 4 \vF_i \cdot \vF_j + 4\epsilon.
\end{align*}

Therefore,
$\vF_i \cdot \vF_j - T(\vF)_i\cdot T(\vF)_j\geq -\epsilon$.
By a similar argument, we have
$\vF_i \cdot \vF_j - T(\vF)_i\cdot T(\vF)_j\leq \epsilon$. Then we have ${\mathcal{A}^-_{ij}}^C\cap {\mathcal{A}^+_{ij}}^C\subset \mathcal{B}_{ij}^C$, and thus
$$
\mathbb{P}(\mathcal{B}_{ij})\leq \mathbb{P}(\mathcal{A}^-_{ij}\cup A^+_{ij})\leq 4\exp\{-\frac{(\epsilon^2-\epsilon^3)k}{4}\}
$$
and
$$\mathbb{P}(\cup_{i<j}\mathcal{B}_{ij})\leq \sum_{i<j}\mathbb{P}(\mathcal{B}_{ij})\leq 4n^2\exp\{-\frac{(\epsilon^4-\epsilon^3)k}{4}\}.$$

This probability is less than $1$ if we take $k> \frac{16 \ln{n}}{\epsilon^2}$. Therefore, there must exist a $T$ such that $\cap_{i<j}\mathcal{B}_{ij}^C$ holds, which completes the proof.   

\textbf{Proof of Proposition 4.1:} Letting $N=WH$, $a_{ij}=(F_1F_1^T)_{ij}$, and $b_{ij}=(F_2F_2^T)_{ij}$ in equation \eqref{eq:ffa_one}, we have:
\begin{multline}
\label{eq:prf_l2_fa}
\mathbb{E}_z \mathcal{L}_{ffa}(F_1,F_2;2) = \mathbb{E}_z \sum_{i=1}^{N}(\sum_{j=1}^{N} {|a_{ij}-b_{ij}|z_j})^2  \nonumber \\
=\mathbb{E}_z \sum_{i=1}^{N} ( \sum_{j=1}^{N} {|a_{ij}-b_{ij}|^2 z_j^2}+2\sum_{j\not= k}|a_{ij}-b_{ij}||a_{ik}-b_{ik}|z_j z_k) \nonumber \\
=\mathbb{E}_z \sum_{i=1}^{N} \sum_{j=1}^{N} {|a_{ij}-b_{ij}|^2 z_j^2}+2\sum_{i=1}^{N}  \sum_{j\not= k}|a_{ij}-b_{ij}||a_{ik}-b_{ik}|z_j z_k \nonumber \\
=\sum_{i=1}^{N} \sum_{j=1}^{N} {|a_{ij}-b_{ij}|^2 \mathbb{E}_z  z_j^2}\\ 
 + 2\sum_{i=1}^{N}  \sum_{j\not= k}|a_{ij}-b_{ij}||a_{ik}-b_{ik}| \mathbb{E}_z  z_j z_k  \nonumber \\
=\sum_{i=1}^{N} \sum_{j=1}^{N} {|a_{ij}-b_{ij}|^2 } = \mathcal{L}_{fa}(F_1,F_2;2).
\end{multline}
\textbf{Proof of Theorem 4.1:}
Given a Gaussian matrix $Z_k=[\mathbf{z}_1, \cdots, \mathbf{z}_k]\in \mathbb{R}^{n\times k}$,
$$\mathcal{L}_{ffa, k}(\Theta) = \frac{1}{k}\sum_{l=1}^k\mathcal{L}_{ffa}(F_1, F_2, \mathbf{z}_l).$$
For any fixed $\Theta$, $\mathcal{L}_{ffa}(F_1, F_2, \mathbf{z}_l)$, $l=1, \cdots, k$, are i.i.d random variables. Suppose the first moment of each random variable is finite, by the strong law of large numbers, $\mathcal{L}_{ffa, k}(\Theta)$ converges to $\mathbb{E}[\mathcal{L}_{ffa}(F_1, F_2, \mathbf{z}_1)]$ almost surely. In other words, $\lim_{k\to \infty}\mathcal{L}_{ffa, k}(\Theta) = \mathcal{L}_{fa}(\Theta)$ with probability 1.
\\
\\
\textbf{Proof of Proposition 4.2:}
 By Chebyshev’s inequality, we have 
 \begin{multline}
\mathbb{P}\big(\big|\mathcal{L}_{ffa, k}(\Theta)-\mathbb{E}[\mathcal{L}_{ffa, k}(\Theta)]\big|>\epsilon\big)\leq\\
\frac{\text{Var}(\mathcal{L}_{ffa, k}(\Theta))}{\epsilon^2}= \frac{\text{Var}(\mathcal{L}_{ffa}(F_1, F_2, \mathbf{z}_1))}{\epsilon^2 k}.
\end{multline}
In order to estimate \begin{multline}
\text{Var}(\mathcal{L}_{ffa}(F_1, F_2, \mathbf{z}_1) = \\\mathbb{E}[\mathcal{L}_{ffa}^2(F_1, F_2, \mathbf{z}_1)] - \big(\mathbb{E}[\mathcal{L}_{ffa}(F_1, F_2, \mathbf{z}_1)]\big)^2,
\end{multline}
it suffices to estimate
$$\mathbb{E}[\mathcal{L}_{ffa}^2(F_1, F_2, \mathbf{z}_1)]=$$
$$\mathbb{E}_z \big(\sum_{i=1}^{N} \sum_{j=1}^{N} {|a_{ij}-b_{ij}|^2 z_j^2}+\sum_{i=1}^{N}  \sum_{j\not= k}|a_{ij}-b_{ij}||a_{ik}-b_{ik}|z_j z_k \nonumber\big)^2$$
which equals (as cross terms are zeros):
$$
\mathbb{E}_z \big(\sum_{i=1}^{N} \sum_{j=1}^{N} {|a_{ij}-b_{ij}|^2 z_j^2}\big)^2$$
$$+\big(\sum_{i=1}^{N}  \sum_{j\not= k}|a_{ij}-b_{ij}||a_{ik}-b_{ik}|z_j z_k \nonumber\big)^2.$$
Direct computation yields:
$$\sum_{i=1}^{N} \sum_{j=1}^{N} {|a_{ij}-b_{ij}|^4 z_j^4}+\sum_{i=1}^{N} \sum_{j=1}^{N}\sum_{l\not= i}^N {|a_{ij}-b_{ij}|^2|a_{lj}-b_{lj}|^2 z_j^4}$$
$$+2\sum_{i=1}^{N} \sum_{j=1}^{N}\sum_{l\not= j}^N {|a_{ij}-b_{ij}|^2|a_{il}-b_{il}|^2 z_j^2z_l^2}$$
$$+\sum_{i=1}^{N} \sum_{j=1}^{N}\sum_{k=1}^N\sum_{l\not= j}^N {|a_{ij}-b_{ij}|^2|a_{kl}-b_{kl}|^2 z_j^2z_l^2}$$
Notice that $\mathbb{E}[z_i^4] = 3$. Taking $\mathbb{E}[\cdot]$, we derive 
the upper bound $3 \|\mathcal{L}_{fa}\|_2^4$.
\medskip 

\textbf{Proof of Theorem 4.2:}
Since $\displaystyle \lim_{k\to \infty}\cL_{ffa, k}(\Theta^*)=\cL_{fa}(\Theta^*)$, it suffices to show that 
$$\lim_{k \to \infty}\ \displaystyle \inf_{\Theta}\cL_{ffa, k}(\Theta) = \cL_{fa}(\Theta^*).$$
Note that
$$\forall \Theta, \lim_{k\to \infty}\cL_{ffa, k}(\Theta)=\cL_{fa}(\Theta)\leq \cL_{fa}(\Theta^*).$$
Then, 
$$\cL_{fa}(\Theta^*)
\geq \lim_{k\to \infty}\inf_{\Theta}L_{ffa, k}(\Theta).$$
On the other hand, for arbitrary $\epsilon>0$, we have:
$$ \exists N \ \ s.t. \ \ \forall k>N \ \ |\cL_{ffa, k}(\Theta)-\cL_{fa}(\Theta)|< \frac{\epsilon}{2}, \, \forall  \Theta$$ 
and there exists a sequence $\{\Theta_k\}$ s.t. 
$$\cL_{ffa, k}(\Theta_k)<\inf_{\Theta}\cL_{ffa, k}(\Theta)+\frac{\epsilon}{2}.$$
Note that $|\cL_{ffa, k}(\Theta_k)-\cL_{fa}(\Theta_k)|<\frac{\epsilon}{2}$ for $k>N$, so: 
$$\cL_{fa}(\Theta^*)-\epsilon\leq \cL_{fa}(\Theta_k)-\epsilon<\inf_{\Theta}\cL_{ffa, k}(\Theta), \ \ \forall k>N.$$
Since $\epsilon$ is arbitrary, taking $k\to \infty$, we have 
$$\cL_{fa}(\Theta^*) \leq \lim_{k\to\infty}\inf_{\Theta}\cL_{ffa, k}(\Theta).$$
\\
\\
\textbf{Proof of Corollary 4.2.1:}
For readability, we shorthand: $\cL_{ffa, k}=f_k$ and $\cL_{fa} = f$. Let $$\mathbf{H}=\frac{\nabla^2 f}{\nabla \Theta \nabla \Theta^T} \succcurlyeq \mathbf{0} \in\mathcal{R}^{n \times n}$$ be the Hessian matrix of FA loss, which is positive semi-definite by convexity of $L_{fa}$. Then, $$\frac{\nabla^2 f_k}{\nabla \Theta \nabla \Theta^T}=Z_k^T\mathbf{H}Z_k \succcurlyeq \mathbf{0}\in \mathbb{R}^{k\times k}$$
which implies the convexity of $f_k$ for all $k$. Moreover, it is clear that $f_k$ is smooth for all $k$ since 
\begin{multline}
\|\nabla f_k(\vx)-\nabla f_k(\mathbf{y})\|= \|Z_k(\nabla f(\vx)-\nabla f(\mathbf{y}))\|\\
\leq L\cdot\|Z_k\|\cdot\|\vx - \mathbf{y}\|.
\end{multline}
We note that $f_k$ is also smooth. Although we cannot claim equi-smoothness since we cannot bound $\|Z_k\|$ uniformly in $k$, the above is sufficient for us to prove the desired result. 
\medskip

For $\forall k$, given any initial parameters $\Theta^0$, by smoothness and convexity of $f_k$, it is well-known that $$\|\Theta_k^t - \Theta_k^* \|\leq \|\Theta^0 - \Theta_k^*\|$$ 
where $\Theta_k^t$ is the parameter we arrive after $t$ steps of gradient descent. Hence, we can pick a compact set $\mathbf{K}= \overline{B_R(\Theta^*)}$ for $R$ large enough such 
that $\{\Theta_k\}_{k=1}^{\infty}\subset\mathbf{K}$ (denote $\Theta^*_{\infty}=\Theta^*$). Now, it's suffices to prove $f_k$ converges to $f$ uniformly on $K$. In fact, $f_k$ converges to $f$ on any compact set. To begin with, we state a known result from functional analysis (\cite{jouak1984equicontinuity, cobzas2017lipschitz}):
\medskip

\begin{lemma}(Uniform boundedness and equi-Lipschitz) 
Let $\mathcal{F}$ be a family of convex function on $\mathbb{R}^n$ and $K\subset \mathbb{R}^n$ be a compact subset. Then, $\mathcal{F}$ is equi-bounded and equi-Lipschitz on $K$.  
\end{lemma}
This result is established in any Banach space in \cite{jouak1984equicontinuity}, so it automatically holds in finite dimensional Euclidean space. By Lemma 6.1, we have that the sequence $\{f_k\}_{k=1}^{\infty}$, where $f_{\infty}=f$, is equi-Lipschitz. $\forall > 0 $, $\exists\, \delta>0$ s.t. $|f_k(x)-f_k(y)|<\epsilon$ for all $k$ and $x, y\in K$ when  $|x-y|<\delta$. Since $\{B(x, \delta)\}_{x\in K}$ forms an open cover for $K$, we have a finite sub-cover $\{B(x_j, \delta)\}_{j=1}^m$ of $K$. Since there are finitely many points $x_j$, there exists $N_{\epsilon}$ such that 
$$\forall k> N_{\epsilon},\ \ |f_k(x_j)-f(x_j)|<\epsilon, \; \text{for }j=1,\cdots, m.$$
For any $x\in K$, $x\in B(x_{j^*}, \delta)$ for some $j^*$. For all $k>N_{\epsilon}$, we have
\begin{multline}
|f_k(x)-f(x)|\leq \\
|f_k(x)-f_k(x_{j^*})|+|f_k(x_{j^*})-f(x_{j^*})|+|f(x_{j^*})-f(x)|\\
\leq (2\Tilde{L}+1)\epsilon
\end{multline}
where $\Tilde{L}$ is the Lipschitz constant for equi-Lipschitz family. Therefore, $f_k$ converges to $f$ uniformly on $K$.\\
\newpage
\bibliographystyle{plain}
\bibliography{egbib}

\end{document}